\documentclass[%
     reprint,
     amsmath,amssymb,
     aps,
    ]{revtex4-2}
    
    \makeatletter
    \renewcommand
    \makeatother
    
    \usepackage{float}
    \usepackage[colorlinks=true, linkcolor=blue, citecolor=blue, urlcolor=blue]{hyperref}
    \usepackage{graphicx}
    \usepackage{dcolumn}
    \usepackage{bm}
    \usepackage[linesnumbered, ruled, vlined]{algorithm2e}

\begin{document}
    
    
    \title{Two-dimensional Parallel Tempering for Constrained Optimization}
    \author{Corentin Delacour}
    \author{M Mahmudul Hasan Sajeeb}
    \author{João P. Hespanha}
    \author{Kerem Y. Camsari}
    \affiliation{
     Department of Electrical and Computer Engineering\\
    University of California, Santa Barbara,  Santa Barbara,
    CA 93106, USA
    }

\begin{abstract}
Sampling Boltzmann probability distributions plays a key role in machine learning and optimization, motivating the design of hardware accelerators such as Ising machines. While the Ising model can in principle encode arbitrary optimization problems, practical implementations are often hindered by soft constraints that either slow down mixing when too strong, or fail to enforce feasibility when too weak. We introduce a two-dimensional extension of the powerful parallel tempering algorithm (PT) that addresses this challenge by adding a second dimension of replicas interpolating the penalty strengths. This scheme ensures constraint satisfaction in the final replicas, analogous to low-energy states at low temperature. The resulting two-dimensional parallel tempering algorithm (2D-PT) improves mixing in heavily constrained replicas and eliminates the need to explicitly tune the penalty strength. In a representative example of graph sparsification with copy constraints, 2D-PT achieves near-ideal mixing, with Kullback-Leibler divergence decaying as $\mathcal{O}(1/t)$. When applied to sparsified Wishart instances, 2D-PT yields orders of magnitude speedup over conventional PT with the same number of replicas. The method applies broadly to constrained Ising problems and can be deployed on existing Ising machines.
\end{abstract}

    \maketitle
    
    \begin{figure*}[t!]
        \centering
        \includegraphics[width=\linewidth]{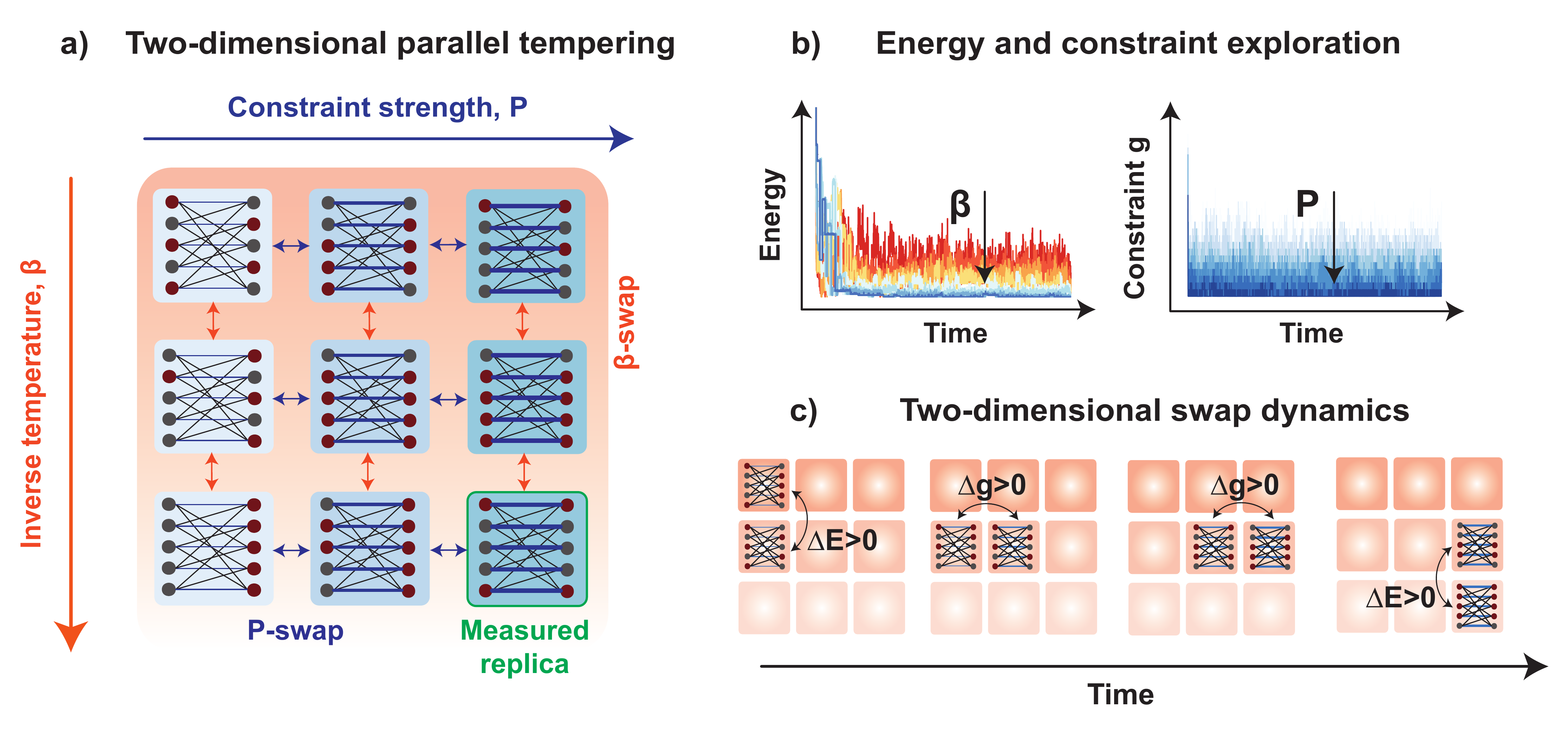}
        \vspace{-20 pt}
        \caption{a) Two-dimensional parallel tempering consists of an $I\times J$ array of replicas. Each row corresponds to a fixed inverse temperature $\beta$, and each column to an increasing constraint strength $P$ weighting the constraint function $g$. We illustrate copy constraints, e.g., $g=(S_1-S_2)^2$ for spins $S_1$ and $S_2$. As in standard parallel tempering, replicas swap along the $\beta$ direction ($\beta$-swaps). Here, we introduce an additional swap dimension along the constraint strength $P$, where replicas at the same $\beta$ exchange states ($P$-swaps). The last column enforces strong constraints and yields high feasibility but mixes slowly. In contrast, early columns mix efficiently but may yield infeasible states. $P$-swaps allow feasible configurations to propagate rightward, enhancing mixing in rigid replicas. b) Energy and constraint trajectories over time for replicas in a fixed column (left) and a fixed row (right), illustrating relaxation along the $\beta$ and $P$ directions, respectively. c) Illustration of 2D swap dynamics. States with lower energy ($\Delta E > 0$) or higher feasibility ($\Delta g > 0$) are preferentially swapped toward the last replica, which accumulates optimal and feasible solutions.
        \vspace{-5pt}}
        \label{fig1}
    \end{figure*}

    \textit{Introduction}---Simulating spin glass systems with the Ising model is a hard problem with  applications beyond physics, from combinatorial optimization \cite{Kirkpatrick_1983} to machine learning \cite{Ackley_1985}. In particular, sampling from the Boltzmann distribution at low temperature is central to optimization but remains challenging for Markov Chain Monte Carlo methods such as Gibbs sampling or simulated annealing, which require a large number of samples for high dimensional problems \cite{aadit_2022}. Recently,  hardware accelerators known as Ising machines have emerged to accelerate optimization and sampling by harnessing new computing architectures and devices \cite{mcmahon_2016,aramon_2019,goto_2019,Honjo_2021,Mohseni_2022,Maher_2024,Graber_2024}.
    
    Despite their success on unconstrained problems, Ising machines are significantly slowed down when constraints are added to the Ising Hamiltonian \cite{sajeeb2025scalable}. Constraints are essential in many real-world applications, such as modeling limited resources in knapsack problems, enforcing task sequences in scheduling \cite{Lucas_2014}, or copying spins to compensate for limited hardware connectivity \cite{venturelli_2015,sajeeb2025scalable}. This slowdown arises because of large energy penalties separating the optimal feasible solutions from infeasible states. The search space can also increase for problems involving auxiliary spins in the energy formulation, like slack variables for inequality constraints \cite{Lucas_2014} or copy nodes in graph sparsification \cite{sajeeb2025scalable}. Reducing the penalty often leads to low-energy but infeasible local minima, making sampling both inefficient and unreliable without careful penalty tuning.

    Sampling constrained energy landscapes can be improved by parallel tempering (PT), or replica exchange Monte Carlo, a powerful algorithm to sample multimodal probability distributions \cite{Swendsen_1986,geyer_1991,Hukushima_1996,Katzgraber2006,bittner_2008}. PT simulates replicas of the same system at different temperatures and improves mixing in the low-temperature replicas by swapping states with high-temperature ones, based on the Metropolis criterion \cite{Metropolis_1953}.
    While PT can accelerate the search for low-energy constrained solutions, it is typically applied to a single energy landscape defined by a fixed penalty strength for the constraints, which must be tuned prior to execution. If the penalty is too weak, the low-energy states are often infeasible; if too strong, feasible states exist but become difficult to reach due to steep energy barriers.

    In this Letter, we extend PT to a second dimension (2D-PT) with multiple energy landscapes to accelerate sampling under constraints. Along this second dimension, the penalty strength modulating constraint violations is gradually increased from soft to hard constraints, with each value corresponding to a column in the 2D array of FIG.~\ref{fig1}a. Analogous to low-temperature replicas in standard PT, the goal is to reduce the relaxation time of heavily penalized replicas (last column) by swapping states from lower-penalty replicas based on a feasibility measure.
    
    2D-PT offers two key advantages compared to PT for constrained problems. First, exploring a family of energy landscapes with increasing penalty strengths eliminates the need to find an optimal penalty value, which itself can be a hard optimization problem \cite{sajeeb2025scalable,venturelli_2015}. Second, feasibility-based swaps along the penalty axis accelerate convergence in hard-constrained replicas while maintaining high feasibility rates. Similar ideas using two-dimensional grids of replicas have been explored in chemical physics with domain-specific constraints \cite{Yan_1999,Gee_2011, Kokubo_2013, Lee_2015, EBRAHIMI_2019}, but their generalization to broader constrained problems is unclear. In particular, general strategies for selecting penalty coefficients are still lacking, and the performance advantage of 2D-PT over classical PT has yet to be demonstrated through direct comparisons.
    
    Building on prior work \cite{Ashtari_2021}, we formalize 2D-PT and derive swap probabilities satisfying detailed balance that are general for any constrained problem. Furthermore, we propose an adaptive algorithm to find temperature values and penalty strengths for any problem, while homogenizing the swap probabilities. This is a crucial step in practice to enhance mixing throughout the array and prevent any exchange bottleneck during replica swaps. Using our adaptive algorithm, we demonstrate 2D-PT's superiority over PT with the example of graph sparsification \cite{sajeeb2025scalable}, which introduces copy constraints between spins, essential for implementing dense graphs on hardware with limited connectivity, such as quantum processors \cite{Arute_2019,king_2023} and Ising machines \cite{aadit_2022}. Compared to running multiple instances of PT with the same number of replicas, 2D-PT achieves a runtime improvement that empirically scales as $\mathcal{O}(N^5)$ for sparsified Wishart instances \cite{Hamze_2020}. While graph sparsification is relevant for near-term Ising hardware, our method applies broadly to constrained optimization problems with arbitrary soft penalties and can accelerate constrained sampling across a wide range of optimization problems.

    \begin{figure*}[t!]
        \centering
        \includegraphics[width=0.9\linewidth]{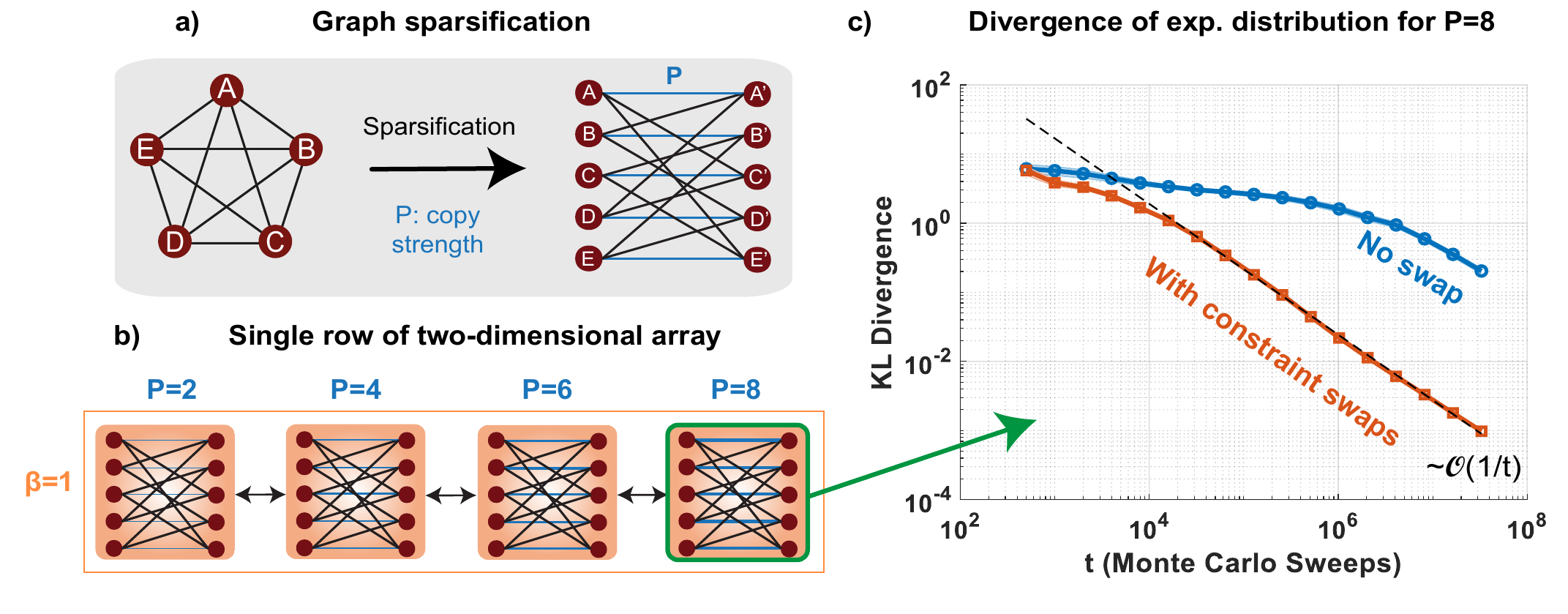}
        \caption{a) Graph sparsification illustrated with a 5-node fully connected graph (e.g., a Full Adder, \cite{sajeeb2025scalable}). Sparsification introduces
          additional copy nodes to reduce node degree, with constraints enforcing identity between copies, expressed as a constraint $g=(S_A-S_{A'})^2=0$. These constraints are embedded in the Ising Hamiltonian via energy penalties  $P\times g$, equivalent to ferromagnetic couplings of strength $P$ between copies. b) To isolate the effect of constraint swaps, we consider a single row of the 2D array at a fixed inverse temperature  $\beta=1$, with 4 replicas at increasing copy strength $P$. c) Kullback-Leibler (KL) divergence between the target Boltzmann distribution (corresponding to the 5-bit Full Adder)  and the empirical distribution from Monte Carlo sampling at $P=8$ with and without $P$-swaps. The empirical distribution is estimated by time-accumulating samples from a single long Markov chain at fixed inverse temperature $\beta = 1$. Without swaps, the high penalty causes slow mixing and high KL divergence. With constraint swaps, feasible configurations are transferred from lower-P replicas, accelerating convergence. The KL divergence scales as $\approx\mathcal{O}(1/t)$, consistent with near-independent sampling (see text). Results are averaged over 100 chains; error bars denote 95\% confidence intervals. The sweep-to-swap ratio is 500.}
        \label{fig2}
    \end{figure*}
    \textit{Two-dimensional parallel tempering}---We extend the parallel tempering algorithm (PT), also called replica exchange Monte Carlo \cite{Swendsen_1986,geyer_1991,Hukushima_1996}, by introducing replicas along a second constraint penalty axis, as shown in FIG.~\ref{fig1}a. Each replica $j$ in this dimension is assigned an energy:
    \begin{equation}\label{eq:E-Pg}
        E_j=f+P_j\,g
    \end{equation}
    where $f$ is the cost function to minimize, $g\geq 0$ is the constraint function with $g=0$ when constraints are satisfied, and $P_j$ is the corresponding penalty strength. 
    
    As the weight $P_j$ increases, the minimum of Eq.~\ref{eq:E-Pg} increases monotonically and converges to that of the constrained optimization \cite{Bertsekas99}.
    This is analogous to the observation that, as the inverse-temperature parameter $\beta$ increases, the support of the Boltzmann distribution converges  to the set of global minima. Not surprisingly, both the increase of $\beta$ and of $P_j$ are detrimental to ground state exploration by introducing steep energy barriers \cite{Lucas_2014}.
    In practice, $P$ is often tuned iteratively to achieve an acceptable fraction of feasible samples \cite{venturelli_2015,sajeeb2025scalable}. 
    
    However, the dual role of $\beta$ and $P_j$ suggests instead extending parallel tempering across two dimensions as a matrix of replicas: each row of the matrix corresponding to one value of the inverse temperature parameter $\beta_i$ and each column to one value of the weight parameter $P_j$, as illustrated in FIG.~\ref{fig1}a. Organizing the rows and columns to reflect increasing values of both parameters, the solution to the constrained optimization is extracted from the bottom-right replica of the matrix. Fast mixing is achieved by coupling this replica with the ones above (lower $\beta$ and thus higher cost) and to the left (lower $P_j$ and thus less feasible). 
    
    This coupling takes the form of
    swap moves along the penalty axis ($P$-swaps), replicated at each temperature, as illustrated in FIG.~\ref{fig1}a.  The acceptance probability for swapping two neighboring replicas in the same row (i.e., at fixed inverse temperature $\beta$) is given by:
    \begin{align}
        P_{swap-P}=\min\big(1,\exp\big(\beta\Delta P \Delta g)\big)
        \label{pswap}
    \end{align}
    where $\beta$ is the inverse temperature of a given row, $\Delta P=P_{j+1}-P_j$, and $\Delta g=g_{j+1}-g_j$, is the energy change in the constraint violation. When a replica at lower penalty strength carries a sample with better feasibility ($\Delta g > 0$), the swap is always accepted, allowing feasible configurations to flow toward the more rigid, high-$P$ replicas. Conversely, swaps that reduce feasibility are accepted with lower probability, especially at large $\beta$. This mechanism promotes mixing among hard-constrained replicas while maintaining high feasibility in the final column.
    The multiplicative $\beta$ indicates that low-temperature replicas will mostly make greedy swaps to increase feasibility, whereas high-temperature replicas accept non-rewarding swaps with higher probability.
    
    As in standard parallel tempering, exchanges also occur along the temperature axis ($\beta$-swaps), independently within each column of the 2D array. The acceptance probability for swapping neighboring replicas in the same column is given by:
    \begin{align}
        P_{swap-\beta}=\min\big(1,\exp\big(\Delta \beta \Delta E)\big)
        \label{beta_swap}
    \end{align}
    where $\Delta \beta =\beta_{i+1}-\beta_i$ and $\Delta E=E_{i+1}-E_i$ is the energy difference between replicas at the two temperatures fixed for penalty $P_j$. These swaps allow low-energy configurations from high-temperature replicas to propagate toward low-temperature replicas, improving mixing and convergence. The exact derivations of these acceptance rules (Eq.~\ref{pswap}-\ref{beta_swap}) 
    are given in Appendix \ref{appendix_derivation} and the full 2D-PT algorithm is outlined in  Appendix \ref{appendix_algs}.
    
    Extending parallel tempering to the penalty strength dimension enables rich swap dynamics,  allowing low-energy and feasible samples propagate towards the target replica at the highest inverse temperature $\beta_I$ and the largest penalty $P_J$, shown at the bottom right in FIG. \ref{fig1}c. This two-dimensional structure contrasts with conventional PT, which operates over a single energy landscape defined by a fixed penalty strength. A natural comparison to our algorithm that uses the same resources is to parallelize PT across J columns but with a fixed penalty strength ($J$-column PT). However, as we show next, the ability to exchange states along the $P-$axis gives 2D-PT a distinct advantage, yielding a dynamic exponent improvement that empirically scales as $\sim \mathcal{O}(N^5)$ in the graph sparsification setting. Moreover, sampling from the last column (with the largest penalty $P_J$) ensures high feasibility, unlike the J-column PT, where feasibility remains limited even when $P$ is optimized.
    
    \begin{figure*}[t!]
        \centering
\vspace{-7.5pt}        \includegraphics[width=0.875\linewidth]{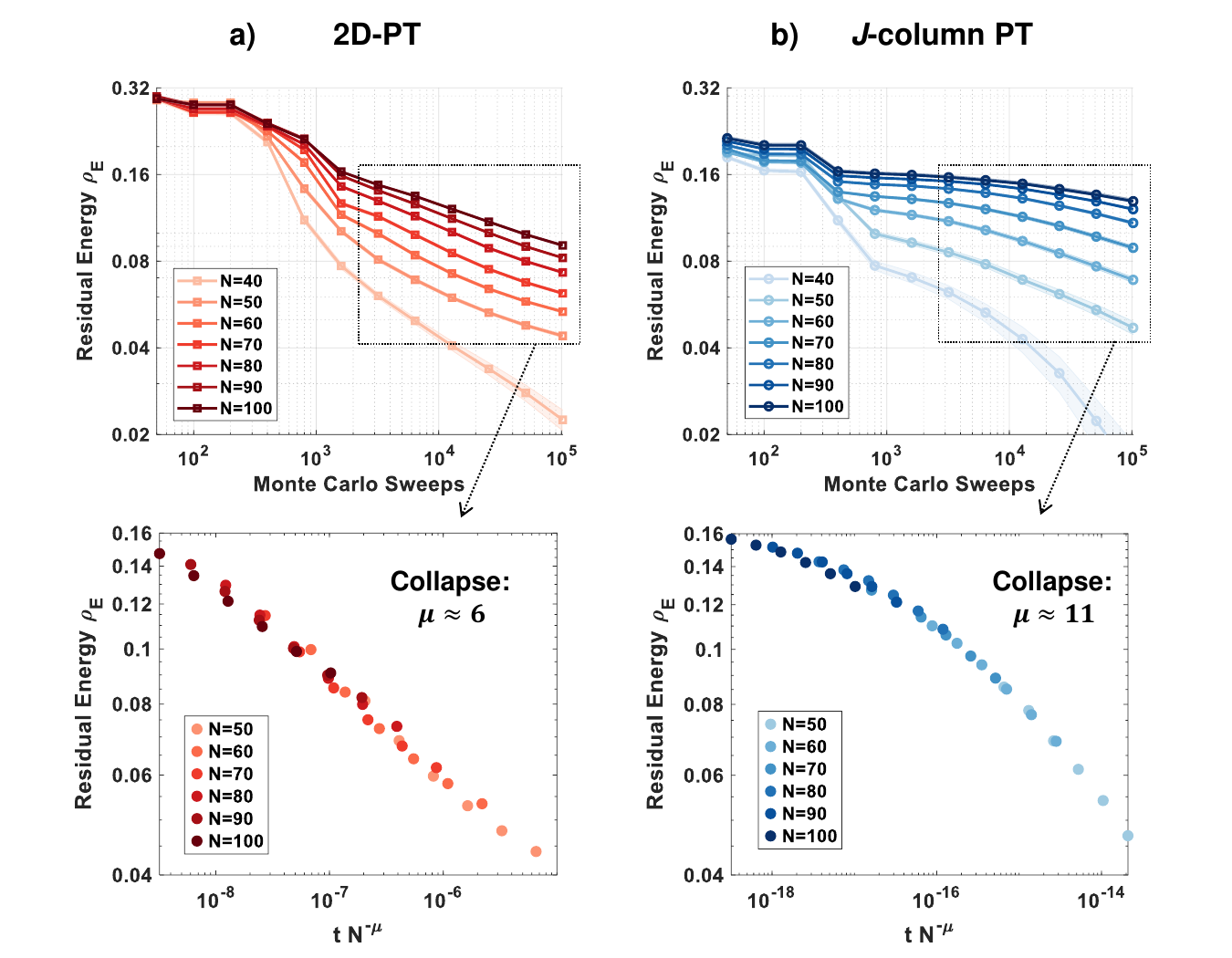}
\vspace{-7.5pt}  
        \caption{a) Comparison between two-dimensional parallel tempering (2D-PT) and standard parallel tempering repeated $J$-times ($J$-column PT) on sparsified planted Wishart instances using three copies per node. To equalize resources, $J$-column PT is run $J$ times with the same $\beta$ schedule and total number of samples as 2D-PT. The penalty strength is set to the mean $P$ value used in 2D-PT, and the best feasible energy is selected post hoc. (a-b) Residual energy $\rho_E = (E - E_{\mathrm{gs}})/N$ as a function of Monte Carlo sweeps $t$, measured in the bottom-right replica of 2D-PT. Each curve is averaged over 200 instances and 10 trials per instance. The sweep-to-swap ratio is 50. Lower panels: finite-size scaling collapse using the ansatz $\rho_E N^b = F(tN^{-\mu})$ with $b = 0$. 2D-PT achieves a collapse with $\mu \approx 6$, while $J$-column PT requires $\mu \approx 11$ over the same time window ($t \geq 3.2 \times 10^3$ MCS), implying a time-to-solution gap scaling as $\mathcal{O}(N^5)$.}
        \label{fig3}
    \end{figure*}

    \textit{Graph sparsification with copy constraints}--- Although the algorithm applies to arbitrary constraint functions with $g \geq 0$, a representative and practically relevant example is the problem of graph sparsification \cite{sajeeb2025scalable}. Many Ising-based hardware accelerators have limited connectivity and require dense problem graphs to be mapped onto sparse physical topologies using additional nodes, or copies. Typically, a compiler has to map each logical node to a chain of physical nodes, distributing its neighbors among them. To preserve the logical problem structure, all copies should assume the same spin value. For instance, two copies $S_1$ and $S_2$ can be constrained via $g=(S_1-S_2)^2=0$. This constraint is equivalent to a ferromagnetic coupling \(-S_1 S_2\), since \(g = (S_1 - S_2)^2 = 2(1 - S_1 S_2)\) in bipolar notation where \(S_i^2 = 1\), and constant offsets in energy are irrelevant.
    
    FIG. \ref{fig2}a shows a 5-node all-to-all graph example, sparsified by introducing two copies per logical node to limit the node degree to 3 neighbors \cite{sajeeb2025scalable}. Instead of optimizing the copy strength $P$, we consider 4 replicas of the resulting 10-node sparse graph with increasing penalty values  $P\in\{2,4,6,8\}$, as shown in FIG. \ref{fig2}b. This configuration corresponds to a single row ($\beta=1$) in the two-dimensional array, isolating the effect of $P$-swaps. To assess sampling accuracy, we compute the Kullback-Leibler divergence (KL) between the empirical distribution of the last replica ($P=8$) and the target Boltzmann distribution of the original 5-node system, as shown in FIG. \ref{fig2}c. The empirical distribution is obtained by time-accumulating samples over a single Markov chain. Without constraint swaps, the large penalty exhibits slows mixing and it takes $4\times 10^6$ Monte Carlo sweeps to reduce the KL divergence below 1. In contrast, enabling $P$-swaps every 500 sweeps yields a two order of magnitude speed-up, with KL falling below 1 in fewer than $2\times 10^4$ sweeps. More importantly, the KL divergence exhibits a scaling of $\mathcal{O}(1/t)$, which matches the theoretical expectation for independent sampling. In the Appendix \ref{appendix_kl}, we derive this behavior analytically and show that the expected KL divergence approaches $(2^N-1)/(2t)$ where $N$ is the number of nodes and $t$ is the Monte Carlo samples. The observed agreement with this ideal scaling confirms that, in this regime, the $P$-swap mechanism yields sampling accuracy comparable to that of i.i.d. samplers.
    
    \textit{Ground-state optimization under constraints}---We now apply the two-dimensional parallel tempering algorithm to all-to-all Ising problems that we sparsify using three copies per node, similar to the earlier two copy example shown in FIG.~\ref{fig2}a. For the problem sizes considered in this work  ($N\leq 100$), three copies per node are typically sufficient to embed dense graphs onto hardware with limited connectivity \cite{sajeeb2025scalable}. Each logical node is mapped to a chain of three physical spins, $S_1$, $S_2$, and $S_3$,  with a copy constraint expressed as  $(S_1-S_2)^2+(S_2-S_3)^2=0$. This condition is satisfied by minimizing the equivalent energy term $-S_1S_2-S_2S_3$. The full constraint function $g$ is the sum over all such local terms and contains $2N$ contributions, each weighted by the penalty strength $P_j$ assigned to column $j$ in the 2D array.
    
    The considered Ising problems are planted Wishart instances with moderate difficulty, controlled by a density parameter $\alpha=0.75$ \cite{Hamze_2020}. For each instance, we employ a custom adaptive procedure (see Appendix \ref{appendix_algs}) to determine the inverse temperatures $\beta_i$ and penalty strengths $P_j$  such that the swap probabilities between adjacent replicas are approximately 0.5 in both  directions. The adaptive algorithm adds rows until the energy standard deviation reaches a preset threshold, and adds columns until the feasibility of the last column approaches 100 \%. The adaptive procedure naturally allocates more replicas in temperature or penalty regions where energy fluctuations peak, often near phase-transition boundaries, thereby maintaining uniform swap rates. Our adaptive treatment of penalty constraints extends feedback-optimized strategies developed for  conventional PT \cite{Katzgraber2006}.  For all numerical experiments, the total number of replicas is capped at $I\times J=20\times 20$ replicas.
    
    After determining the $\beta_i$ and $P_j$ values, we run 2D-PT with a sweep-to-swap ratio of 50 and evaluate  the residual energy per spin $\rho_E=(E_{I,J}-E_{\mathrm{gs}})/N$, where $E_{I,J}$ is the energy of the bottom-right replica, corresponding to the  highest inverse temperature $\beta_I$ and the strongest penalty strength $P_J$. This replica is expected to yield low-energy and fully feasible configurations ($g\approx0$). FIG.~\ref{fig3}a shows $\rho_E$ as a function of Monte Carlo sweeps for increasing logical problem sizes $N\leq 100$ (corresponding up to 300 physical spins). Each data point is averaged over 200 instances and 10 independent trials per instance. After an initial transient time of approximately $10^3$ sweeps,  2D-PT  reduces the residual energy with a power-law trend over time for the range of $\rho_E$ considered.
    
    To quantify how the relaxation time grows with problem size we adopt the finite-size-scaling (FSS) ansatz
\begin{equation}
  \rho_E\,N^{\,b}=F\!\bigl(t\,N^{-\mu}\bigr),
  \label{FFS}
\end{equation}
where \(F\) is a universal scaling function, \(\mu\) is the dynamic exponent, and \(b\) absorbs any residual size dependence \cite{huse1989}. For the planted Wishart ensemble the total energy is extensive in the number of
logical spins~\cite{Hamze_2020}; sparsification redistributes the couplers
without altering that leading \(N\) scaling. Assuming therefore that the residual density
\(\rho_E=(E-E_{\mathrm{gs}})/N\) remains \(\mathcal O(1)\), we set \(b=0\).

FIG.~\ref{fig3}a shows the resulting collapse of the energy residuals for $t\geq 3.2\times 10^3$ Monte Carlo sweeps obtained with a of $\mu\approx 6$. This exponent exceeds the $\mu\approx 4$ reported  for  sparsified Erdős–Rényi  graphs with three copies \cite{sajeeb2025scalable}, indicating that the
planted Wishart instances present a tougher dynamical landscape. 
    
    We now compare  2D-PT  with a baseline strategy in which standard PT is repeated $J$ times at a fixed penalty strength, chosen to match the mean value of the penalties used in 2D-PT ($J$-column PT).  Each instance uses the same \(\beta\) schedule, and the best feasible energy among the \(J\) runs is selected.  While this approach uses the same number of total replicas and samples as 2D-PT, feasibility is not guaranteed and must be verified post hoc, which can incur overhead, especially for general constraint types.  For the copy constraint problem considered here, the selected $P$ yields  feasible configurations with probability  between 20\% and 60\%.
    
    As shown in FIG.~\ref{fig3}b, $J$-column PT reduces the residual energy at a significantly slower rate than 2D-PT. Applying the same FSS analysis to the $J$-column data yields a dynamic exponent of $\mu\approx 11$ compared to $\mu\approx 6$ for 2D-PT. Under the scaling form of Eq.~\ref{FFS}, this implies that the time required to reach a target residual $\rho_0$ grows as  $\mathcal{O}(N^5)$ faster in $J$-column PT. For example, reaching a target residual $\rho_0\approx0.05$ at $N=200$ (600 physical spins) requires   $1.3\times 10^8$   sweeps with 2D-PT, compared to $2\times 10^{11}$ for $J$-column PT. The ability of 2D-PT to transfer low-energy, high-feasibility configurations along the penalty axis accounts for this improvement.  While the precise scaling  gain may vary across  problem classes, these results   stress the potential of 2D-PT to accelerate constrained optimization beyond what can be achieved with conventional parallel tempering.

\textit{Conclusion}---We introduced a two-dimensional extension of parallel tempering (2D-PT) for constrained Ising problems, where a second replica axis with increasing penalty strengths enables interpolation from soft to hard constraints. Feasibility-based swaps along this axis allow low-penalty replicas to transfer configurations to high-penalty ones, accelerating convergence without the need for penalty tuning. In simple but representative examples, 2D-PT achieves near-optimal sampling accuracy, with Kullback–Leibler divergence decaying as \(\mathcal{O}(1/t)\), consistent with independent sampling. Applied to sparsified planted Wishart instances, 2D-PT reduces residual energy significantly faster than a baseline with matched resources. Finite-size scaling reveals a dynamic exponent improvement of \(\Delta \mu \approx 5\), corresponding to an \(\mathcal{O}(N^5)\) time-to-solution gap. 2D-PT computes constraints only during swaps and naturally concentrates feasible samples in its final column, making it compatible with existing Ising hardware and broadly applicable to constrained optimization and constrained sampling problems.
    
  \textit{Acknowledgments}---Authors acknowledge support
    from the Office of Naval Research (ONR), Multidisciplinary University Research Initiative (MURI) grant
    N000142312708.

    \bibliography{mybib}

     \appendix 
    
    \section{Swap probabilities} \label{appendix_derivation}
    \subsection{General case}
    Consider two replicas A and B sampling independently from their respective Boltzmann distributions at inverse temperatures $\beta_A$, $\beta_B$:
    \begin{equation}
        \pi_A = \frac{\exp\big[-\beta_A E_A(S_A)\big]}{Z_A},\quad \pi_B = \frac{\exp\big[-\beta_B E_B(S_B)\big]}{Z_B}
    \end{equation}
    where $Z_A$, $Z_B$ are the partition functions. The joint probability assumes that A and B sample independently from their respective Boltzmann distributions:
    \begin{align}
        &\pi\big(E_A(S_A),E_B(S_B)\big)= \\ \nonumber
        &\frac{1}{Z_A Z_B}\exp\big[-\beta_A E_A(S_A)-\beta_B E_B(S_B)\big]
    \end{align}
   A swap exchanges the states of replicas A and B with probability \(P(E_A(S_A), E_B(S_B))\).

    Detailed balance requires:
    \begin{align}
        &\pi\big(E_A(S_A),E_B(S_B)\big) P\big(E_A(S_A),E_B(S_B)\big)\\ \nonumber
        =&\pi\big(E_A(S_B),E_B(S_A)\big) P\big(E_A(S_B),E_B(S_A)\big)
    \end{align}
    Substituting the joint distribution to the detailed balance condition yields: 
    \begin{align}
        \frac{P\big(E_A(S_A),E_B(S_B)\big)}{P\big(E_A(S_B),E_B(S_A)\big)}=\exp\big[\beta_A \Delta E_A+\beta_B \Delta E_B\big] \label{swap_def}
    \end{align}
    where $\Delta E_A=E_A(S_A)-E_A(S_B)$ and $\Delta E_B=E_B(S_B)-E_B(S_A)$ are the energy differences when exchanging the replicas states. Applying the Metropolis criterion to satisfy this condition, gives the swap probability:
    \begin{align}
        P_{swap}&=P\big(E_A(S_A),E_B(S_B)\big)\\ \nonumber
        &=\min\big(1,\exp\big[\beta_A \Delta E_A+\beta_B \Delta E_B\big]\big)
    \end{align}
    
    \subsection{Special case of constrained optimization}
    
    We now consider constrained optimization problems where constraints are modeled by a scalar function $g(S)\geq0$ and added to the cost function $f(S)$ as:
    \begin{equation}
        E_A(S)=f(S)+P_A\,g(S)
    \end{equation}
    with $P_A>0$ a penalty parameter penalizing the energy of replica A when $g(S)>0$. In this context, the energy differences for the swap defined in Eq. \ref{swap_def} become:
    \begin{align}
        \Delta E_A    &= \Delta f+P_A \, \Delta g \\ \nonumber
        \Delta E_B    &= -\Delta f-P_B \, \Delta g
    \end{align}
    where $\Delta f=f(S_A)-f(S_B)$ and $\Delta g=g(S_A)-g(S_B)$. Finally, the exponential term from the swap probability is expressed as:
    \begin{align}
        &\exp\big[\beta_A \Delta E_A+\beta_B \Delta E_B\big]\\ \nonumber
        =&\exp\big[\Delta f\,(\beta_A-\beta_B)+\Delta g\,(\beta_A P_A-\beta_B P_B)]
    \end{align}
    
    We propose to explore constrained solutions along the $P$-direction at fixed $\beta=\beta_A=\beta_B$ corresponding to a fixed row of the 2D array. Denoting $\Delta P=P_A-P_B$, we express the swap probability in the $P$-direction as:
    \begin{align}
        P_{swap-P}=\min\big(1,\exp\big[\beta\Delta P \Delta g]\big)
    \end{align}
  
    Similar to standard PT for a fixed $P$-value, we express the swap probability in the $\beta$-direction as:
    \begin{align}
        P_{swap-\beta}=\min\big(1,\exp\big[\Delta \beta \Delta E]\big)
    \end{align}
    where $\Delta \beta=\beta_A-\beta_B$ between two replicas within a column of the 2D array. \vspace{5pt}

    \section{Bias of the empirical KL divergence}
    \label{appendix_kl}
    
    We outline the standard bias expansion of the empirical Kullback–Leibler (KL) divergence.  
    Let \(p=\{p(i)\}_{i=1}^k\) be a fixed discrete distribution and
    \(\hat p_t(i)= t^{-1}\sum_{s=1}^{t}\mathbf 1_{\{S_s=i\}}\) the histogram from \(t\) i.i.d.\ samples.
    Write \(\delta_i=\hat p_t(i)-p(i)\); then
    \(\mathbb E[\delta_i]=0\) and \(\mathrm{Var}[\delta_i]=p(i)(1-p(i))/t\). For small $\delta_i$, we have: 
    \begin{align}
    \log\!\frac{\hat p_t(i)}{p(i)}
       &= \frac{\delta_i}{p(i)} - \frac{\delta_i^{2}}{2p(i)^{2}} + \cdots \\[6pt]
    D_{\mathrm{KL}}(\hat p_t\Vert p)
       &= \sum_{i}\bigl[p(i)+\delta_i\bigr]
          \Bigl(\frac{\delta_i}{p(i)} - \frac{\delta_i^{2}}{2p(i)^{2}}\Bigr)
          + \cdots 
    \end{align}
    The linear term vanishes because \(\sum_i\delta_i=0\); to leading order
    \[
    D_{\mathrm{KL}}(\hat p_t\Vert p)=\tfrac12\sum_{i}\delta_i^{2}/p(i)+\ldots
    \]
    and hence
    \[
    \mathbb E[D_{\mathrm{KL}}]=\frac12\sum_{i}\frac{\mathrm{Var}[\delta_i]}{p(i)}
    =\frac{1}{2t}\sum_{i}(1-p(i))
    =\frac{k-1}{2t}+\mathcal O(t^{-2})
    \]
    in agreement with Eq. 4.5-4.6 of Paninski \cite{paninski2003}.  
    For correlated samples the same calculation with \(t\to t/\tau_{\mathrm{int}}\) gives
    \(\mathbb E[D_{\mathrm{KL}}]\simeq (k-1)\tau_{\mathrm{int}}/(2t)\). 
    
    \onecolumngrid
    
    \section{Algorithms} \label{appendix_algs}
    
    \begin{algorithm}[H]
    \caption{Two-dimensional parallel tempering}
    \label{alg}
    \KwIn{Ising coefficients $J, h$, function $g$, ordered vectors $\beta$ and $P$, number of MCS, MCS\_per\_swap, initial spins $S_{0}$.}
    \KwOut{Stored spins from the last replica}
    
    $N_{swap} \gets \lfloor MCS / MCS\_per\_swap \rfloor$ \;
    $S_{temp} \gets S_{0}$ \;
    $swap\_direction \gets 1$ \;
    
    \For{$n= 1$ \textbf{to} $N_{swap}$}{
        $even\_swap \gets (n \mod 2 == 0)$ \;
    
        \For{each replica}{
            Perform $MCS\_per\_swap$ sweeps ($S_{temp}$) \;
            For final sweep: compute energy $E$ and  constraint $g$ \;

        }
        
        \
        
        \tcc{Swaps along $P$ direction}
        \If{$swap\_direction = 1$}{
            \For{each $\beta$}{
               \For{$p= 1 + even\_swap:2:length(P) - 1$}{
                    $\Delta P \gets P(p+1)-P(p)$ \;
                    $\Delta g \gets g(p+1)-g(p)$ \;
                    $P_{swap} \gets \min\big(1,\exp(\beta \Delta P \Delta g)\big)$ \;
                    \If{rand $\leq P_{swap}$}{
                        Swap configurations between $p+1$ and $p$ in $S_{temp}$ \;
                    }
                }
            }
        }
        \tcc{Swaps along $\beta$ direction}
        \If{$swap\_direction = -1$}{
            \For{each $P$}{
                \For{$b=1 + even\_swap:2:length(\beta) - 1$}{
                    $\Delta \beta \gets \beta(b+1)-\beta(b)$ \;
                    $\Delta E \gets E(b+1)-E(b)$ \;
                    $P_{swap} \gets \min\big(1,\exp(\Delta \beta \Delta E)\big)$ \;
                    \If{rand $\leq P_{swap}$}{
                        Swap configurations between $b+1$ and $b$ in $S_{temp}$ \;
                    }
                }
            }
        }
    
        Store spins from the last replica \;
    
        \If{$even\_swap$}{
            $swap\_direction \gets -swap\_direction$ \;
        }
    }
    
    \Return Stored spins from last replica \;
    
    \end{algorithm}
    
    \begin{algorithm}[h]
    \caption{Adaptive schedule for two-dimensional parallel tempering}
    \label{alg_APT}
    \KwIn{Initial $\beta_{0}$ and $P_{0}$, maximum number of $\beta$ values $I$, maximum number of $P$ values $J$, minimum energy standard deviation $\sigma_{min}$, learning rates $\alpha_\beta$, $\alpha_P$, $N_{chain}$, number of MCS, Ising coefficients $J$, $h$.}
    \KwOut{$\beta$ and $P$ vectors}
    
    \
    
    \tcc{Adaptive schedule on first column $P_0$:}
    $j\gets 1$;
    $i\gets 1$ \;
    $\beta_s(i,j) \gets \beta_0$;
    $P_s(i,j) \gets P_0$ \;
    
    \While{$i\leq I$}{ 
        \tcc{Sampling population:}
        Initialize $N_{chain}$ Markov chains\;
        \For{$c = 1$ \textbf{to} $N_{\text{chain}}$}{
           Perform MCS sweeps for replica ($\beta_s(i,j),P_0$) \;
           Compute MCS energies $E$ and constraints $g$ \;
        }
    
        Compute population averages for $\sigma_E$ and $\sigma_g$ \;
    
        \
        \If{$\sigma_E>\sigma_{min}$}
            {$\beta_{s}(i+1,j) \gets \beta_{s}(i,j) + \alpha_{\beta}/\sigma_E$ \;
            $P_{s}(i,j+1) \gets P_{s}(i,j) + \alpha_{P}/\big(\beta_s(i,j) \sigma_g\big)$ \;
            $i\gets i+1 \;$}
            \Else{\textbf{break};}
            
    }
    $I \gets i$ \;
    $P(j+1) \gets \text{median}(P_{s}(:,j+1))$ \;
    
    \ 
    
    \tcc{Adaptive schedule for remaining columns:}
    $j \gets 2$; $i \gets 1$ \;
    $\beta_s(i,j) \gets \beta_0$; $P_s(i,j) \gets P(j)$ \;
    \While{$j \leq J$}{

        \For{$i = 1$ \textbf{to} $I$}{ 
            Sample population of $N_{chain}$ for replica ($\beta_s(i,j),P(j)$)\;    
            Compute population averages for $\sigma_E$, $\sigma_g$, and $mean\_g$ \;
    
            \
            
            $\beta_{s}(i+1,j) \gets \beta_{s}(i,j) + \alpha_{\beta}/\sigma_E$ \;
            
            $P_{s}(i,j+1) \gets P_{s}(i,j) + \alpha_{P}/\big(\beta_s(i,j) \sigma_g\big)$ \;
        }
        
        \If{$mean\_g < 0.5$}{  \tcp{Cold replica satisfies constraints}
            \textbf{break}\;
        }
        \Else{
            
            $P(j+1) \gets \text{median}(P_{s}(:,j+1))$ \;
           
            $j \gets j + 1$\;
        }
    }
    
    \tcc{Final $\beta$ schedule:}
    $\beta \gets \text{median}(\beta_{s})$ in column direction\;
    
    \Return{$\beta$, $P$}\;
    
    \end{algorithm}
    
    \end{document}